\documentclass{article}

    \usepackage[preprint, nonatbib]{neurips_2020}

\usepackage[utf8]{inputenc} 
\usepackage[T1]{fontenc}    
\usepackage{hyperref}       
\usepackage{url}            
\usepackage{booktabs}       
\usepackage{amsfonts}       
\usepackage{nicefrac}       
\usepackage{microtype}      

\title{TransformNet: Self-supervised representation learning through predicting geometric transformations}

\author{%
  Sayed Hashim${}^{*}$ \qquad Muhammad Ali\thanks{First authors}\\
  Mohamed Bin Zayed University of Artificial Intelligence, UAE \\
  \texttt{\{sayed.hashim, muhammad.ali\}@mbzuai.ac.ae} \\
    \\
}

\begin{document}

\maketitle

\begin{abstract}
Deep neural networks need a big amount of training data, while in the real world there is a scarcity of data available for training purposes. To resolve this issue unsupervised methods are used for training with limited data. In this report, we describe the unsupervised semantic feature learning approach for recognition of the geometric transformation applied to the input data. The basic concept of our approach is that if someone is unaware of the objects in the images, he/she would not be able to quantitatively predict the geometric transformation that was applied to them. This self supervised  scheme is based on pretext task and the downstream task. The pretext classification task to quantify the geometric transformations should force the CNN to learn high-level salient features of objects useful for image classification. In our baseline model, we define image rotations by multiples of 90 degrees. The CNN trained on this pretext task will be used for the classification of images in the CIFAR-10 dataset as a downstream task. we run the baseline method using various models, including  ResNet, DenseNet, VGG-16, and NIN with a varied number of rotations in feature extracting and fine-tuning settings. In extension of this baseline model we experiment with transformations other than rotation in pretext task. We compare performance of selected models in various settings with different transformations applied to images,various data augmentation techniques as well as using different optimizers. 
This series of different type of experiments will help us demonstrate the recognition accuracy of our self-supervised model when applied to a downstream task of classification.\\

\textbf{Code: }\url{https://github.com/hashimsayed0/TransformNet}
\end{abstract}

\section{Introduction}
Deep neural networks, especially convolutional neural
networks (CNNs) lead to big accomplishments in the computer vision field.
If manually labelled data like ImageNet is available, we can train CNNs models by using back propagation. As a result, CNN's give a state-of-the-art performance on many tasks like image classification and object detection \cite{rotation}.
For real-world problems, a very limited amount of data is available for training purposes, so expensive efforts concerning time and resources are required to provide these labelled training data. This problem leads to a big increase in the interest of researchers to learn deep feature representations in an unsupervised fashion to solve emerging visual understanding tasks with insufficient labelled data \cite{aet}. 

Self-supervised learning technique can be scaled to real-world problems and it provides a reliable methodology for learning visual representations \cite{rotation}. 
In the first step the underlying structural information of data as well as pretext tasks is utilized for feature extraction which can be used to train neural networks[1]. Pretext tasks helps in creating supervisory signals without much expensive efforts. In the second step as downstream tasks existing data sets of images can be used for model training[3].
We believe that in order for a  model to successfully  predict the rotations, it needs to understand the true concept of object by its various properties including its position, pose and location. 
Our work describes a self-supervised learning approach and we suggest image representations learning by training CNN's to recognize the geometric transformation when applied to an input image.
We propose to learn high level semantic image features by training CNNs to recognize the geometric transformations applied to them. 
As the pretext task, we train the network to predict the degrees of rotation and shear applied to the unlabelled input images. 

The pretext classification task to quantify the geometric transformations applied on images should force CNN to learn high-level salient features of objects useful for image classification \cite{rotation}.
These features may include location of objects in the image, type of objects and their orientation in the image. 
After learning the orientation in the image, model relates this orientation to the actual transformation applied  on the image. So this technique proves to be simple yet very power  technique  which provides supervisory signal in SSL-domain. 

We believe that working in this domain which  purely focuses on geometric transformation based self-supervised techniques is necessary due to 2 reasons. (1) Geometric transformations have proved to be simple yet powerful supervisory signals in unsupervised representation learning. (2) Many successful works have used geometric transformations from different paradigms such as autoencoding and classification. To enable a detailed and in-depth understanding of SSL with geometric transformations we can experiment with various state of the art models using various tricks like data augmentation to improve variability of data.

For these reasons, we start with baseline model. In the baseline model, we define transformations as the image rotations by multiples of 90 degrees. The CNN trained on this pretext task is used for object recognition in images in the CIFAR-10 dataset as a downstream task. This helps in measuring the accuracy of our self-supervised model when applied to a particular task of classification. After successfully implementing the baseline models where pretext task is comprised of rotation prediction by network we extend the model in two domains. 

In one domain we train the models using various levels of augmentations.
These augmentations are applied on top of rotations to improve the  generalization ability of the model along with inherent advantage of enhancement in diversified data. Performance of model learning in pretext task using this augmented data is tested in downstream tasks in unfrozen as well as frozen settings to get a fair comparison.

Second domain is the architectural selection and hyper parameters tuning: In this domain we test different  models to compare the results quantitatively and qualitatively. These models include VGG16, NIN, DensNet, ResNetV1 and ResnetV2.Comparison  of these models with  different depth level of convolution blocks i.e 1,2,3,4 and 5 give us better understanding of the performance. In order to thoroughly evaluate and analayse the performance of our model we conduct series of experiments to predict the transformations applied to image.
This series of different type of experiments will help us demonstrate the recognition accuracy of our self-supervised model when applied to a downstream task of classification.

We organize the rest of the paper as follows. Section 2 shows work distribution and contribution. In section 3 we review the related work along with a brief comparison of our work. In section 4 we briefly discuss data exploration and prepossessing. We present our suggested model  in section 5 describing in detail the methodology adopted. we describe the experiments  in section 6  using different models with reference to data augmentation techniques,network depth level,hyper-parameters tuning and optimizer selection. Section 7 contains information about results and data sets. In this section we present quantitative and qualitative outcomes for better analysis and conclusion. Section 8 concludes with brief summary of our work.



\section{Related works}
\subsection{Self supervised learning}
In some methods of self supervisory learning, the supervisory signal is generated from the data itself e.g in a few methods part of data is recovered in pretext task and the network learns during the process \cite{decoupling}. Examples of a few such tasks include image completion \cite{image_completion}, image colourization \cite{image_colorization} and channel prediction \cite{channel_prediction}. Other types of the technique include those which takes underlying concept information of the image to ultimately match the constraints. These techniques include counting \cite{counting}, rotation \cite{rotation} and instance discrimination \cite{instance_disc}.

One of the works propose an idea to train neural networks by using the prediction of the relative positions of two randomly sampled patches \cite{patch_position}. In one other experiment, the concept of the jigsaw puzzle solution by the network is used  for training a convolutional neural network \cite{jigsaw_puzzle}. Another work introduces auto-encoding transformations to exploit the representation of features after different transformations \cite{aet}. To sum it up, these methods help in training the network using self-supervised objectives and thus reduces the dependency on manually labelled data.


Many classical hand-crafted features including SIFT \cite{sift} and RIFT  \cite{rift} for computer vision show insensitivity to some types of transformations. Data augmentation can be used for transformation invariance. We  use the data augmentations in the final phase of our  project, where we use  two types of transformation sets ;10 transformation set and 5 transformation set  in unsupervised task. These data augmentations not only help increase the data but also it increase the generalizability of the model .This further  helped us obtain rotational invariance against transformations.

\subsection{Feature Decoupling}
Some authors proposed the idea of using rotation feature decoupling in self-supervised learning \cite{decoupling}. In this method, instead of working on a pretext task, the main theme is to focus on the attributes of the learned representations by the network as well as their generalization ability. This helps to improve performance in those tasks where we need rotation invariance. The novel scheme is presented which comprised of two tasks, rotation prediction and individual instance discrimination tasks. After predicting image rotation tasks on our baseline model in a midterm, we utilize different types of augmentations in the final phase of our project. 

\subsection{Auto-Encoders}

The basic methodology relies on a assumption that in order to rebuild the data feature representation must have enough information. 
Huge variety of encoders have been proposed including, the variations auto-encoder which explicitly introduces probabilistic assumption about distribution of features extracted from data. Denoising auto-encoder [27]  initiates learning more robust representation by reconstructing original inputs from noise-corrupted inputs.
In this technique  a novel methodology of  unsupervised representation learning using Auto-Encoding Transformation (AET) is used as against the Auto-Encoding Data (AED)approach.  In this technique after application of random transformation, AET tries to predict it using encoded features as accurately as possible at the output end. The central idea in this scheme is that as long as the unsupervised features can encode the fundamental information about original and transformed images, network can predict the transformation.


\section{Data exploration and preprocessing}
\textbf{Dataset}: We used CIFAR-10 dataset \cite{cifar} for our pretext rotation prediction task and downstream object recognition task. The CIFAR-10 dataset contains 60,000 32x32 colour pictures belonging to ten classes, with 6000 pictures in one class. It consists of 50,000 training images and 10,000 test images. 
The test batch comprises of 1000 images from each class. The training batches hold the rest of the images in random order. The training batches contain 5000 images from each class.

\textbf{Preprocessing}: To obtain the data set for pretext task, we constructed data sets with relevance to our experiments. In phase I of our project, we applied rotation to each image in CIFAR-10 dataset and obtained 120,000, 240,000 and 480,000 images in the case of 2, 4 and 8 rotations respectively. In phase II, we applied geometric transformations to each image, resulting in 600,000 and 300,000 images, for 10 and 5 transformations respectively. 

In phase II, we performed data preprocessing in 2 ways. (1) In the first method, we applied each transformation separately to each image, i.e., each transformed image contained only one of the four transformations, rotation, shearing, scaling and translation. (2) In the second method, we applied transformations randomly to each image, resulting in 5 or 10 transformed copies of each image. This means multiple transformations mentioned above could be applied on the same image. We found that first method worked better, so we went ahead with it.

\section{Methodology}
\subsection{Overview}
This work aims to learn meaningful features based on CNN in an unsupervised way. For that purpose, a CNN model $F(.|)$ is trained to quantitatively predict the geometric transformation that are applied to images that the network receives as input. \\

Precisely, consider a set of $K$ geometric transformations $T$ defined as $T = \{t(.|y)\}_{y=1}^{K}$ where $t(.|y)$ is a geometric transformation applied to an image and results in a transformed image $X^{y}$ given as $X^y = t(X|y)$. The CNN learns a function $F(.|y)$ that can predict the geometric transformation $t(.|y)$ that is applied to the input. \\

Here, provided a set of N images, the training objective that the CNN learns to solve is: 
\begin{equation}
    \min _{\theta} \frac{1}{N} \sum_{i=1}^{N} loss\left(X_{i}, \theta\right)
\end{equation}

whereby $loss$ is defined as:
\begin{equation}
    loss\left(X_{i}, \theta\right)=-\frac{1}{K} \sum_{y=1}^{K} \log \left(F^{y}\left(t\left(X_{i} \mid y\right) \mid \theta\right)\right)
\end{equation}



\subsection{Rotation}
The geometric transformations $T$ must force the CNN to learn semantic features. In the process of learning to predict the geometric transformations applied to the image, the CNN would also have to learn the features in the image. Therefore, to get better at this task, the CNN is forced to learn the features of the images. Precisely, to be able to predict the rotation applied to an image, the CNN model should learn to localize the subjects and objects in the image, identify their orientation, direction and type of object, and then connect the orientation of the object with the main orientation shown by different types of object within the images \cite{rotation}.

\begin{figure}
        \caption{Figure depicting the proposed self supervised task for meaningful learning of features \cite{rotation}. Given two plausible geometric transformations, rotations of zero and 180 degrees, a CNN model F(.) is trained to identify the rotation applied to the image that it receives as input. $F^y (X^y)$ is the prediction by model $F(.)$ that states the probability of rotation $y$ when it receives as input an image which was transformed by the rotation $y$}
        \centering
        \includegraphics[width=1\textwidth]{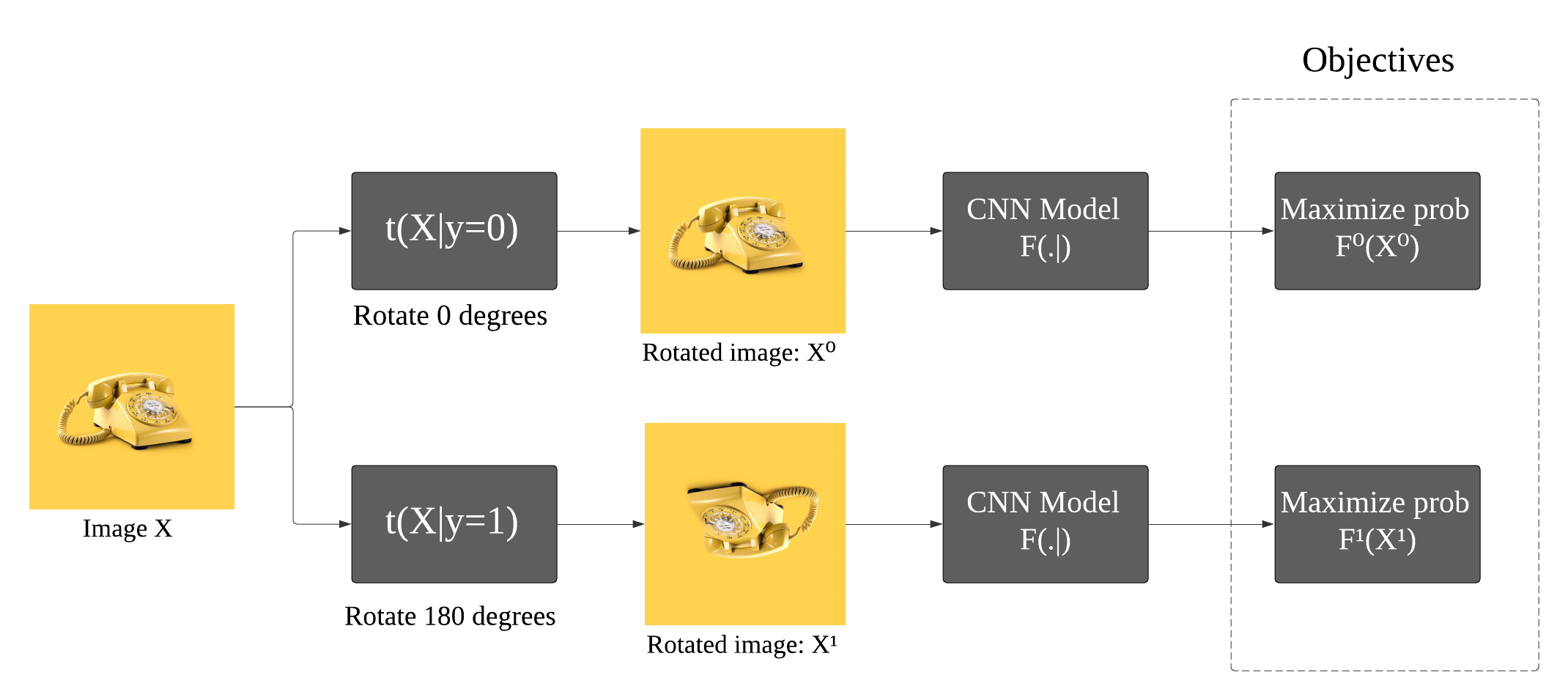}
        \label{fig:illu}
\end{figure}

One such geometric transformation is rotation \cite{rotation}. To predict the amount of rotation that is applied to an image, the CNN would have to learn the semantic features of this image. In this way, rotation provides a powerful supervisory signal. In this work, we choose $T$ as a set of rotations applied to an image, as shown in Figure \ref{fig:illu}. We experimented with different set of rotations as described in the experiments section, and found that $T = \{0$\degree$, 180$\degree$\}$ provides best results. In our upcoming work, we will experiment with more geometric transformations.

A benefit of applying rotations by multiples of ninety degrees instead of other geometric transformations, is that rotations can be applied using flip as well as transpose operations that don't expose any easily identifiable low-level visual artifacts that can help the CNN learn easy features that have no practical value for the visual perception tasks \cite{rotation}. For example, choosing scale as transformation would provide the CNN with clues about the scale factor without having to learn any semantic features in the image.

Moreover, images captured by humans usually are in up-standing position, which give no ambiguity for the rotation task except for round objects.

The above mentioned reasons show that rotation is a well defined simple yet powerful task for self supervision. 

\subsection{Other affine transformations}
\begin{figure}
        \caption{Figure depicting the proposed self supervised task for meaningful learning of features. After applying various transformations to input image $X$ and obtaining $t(X)$, a CNN model is trained to identify the transformations applied to the image}
        \centering
        \includegraphics[width=1\textwidth]{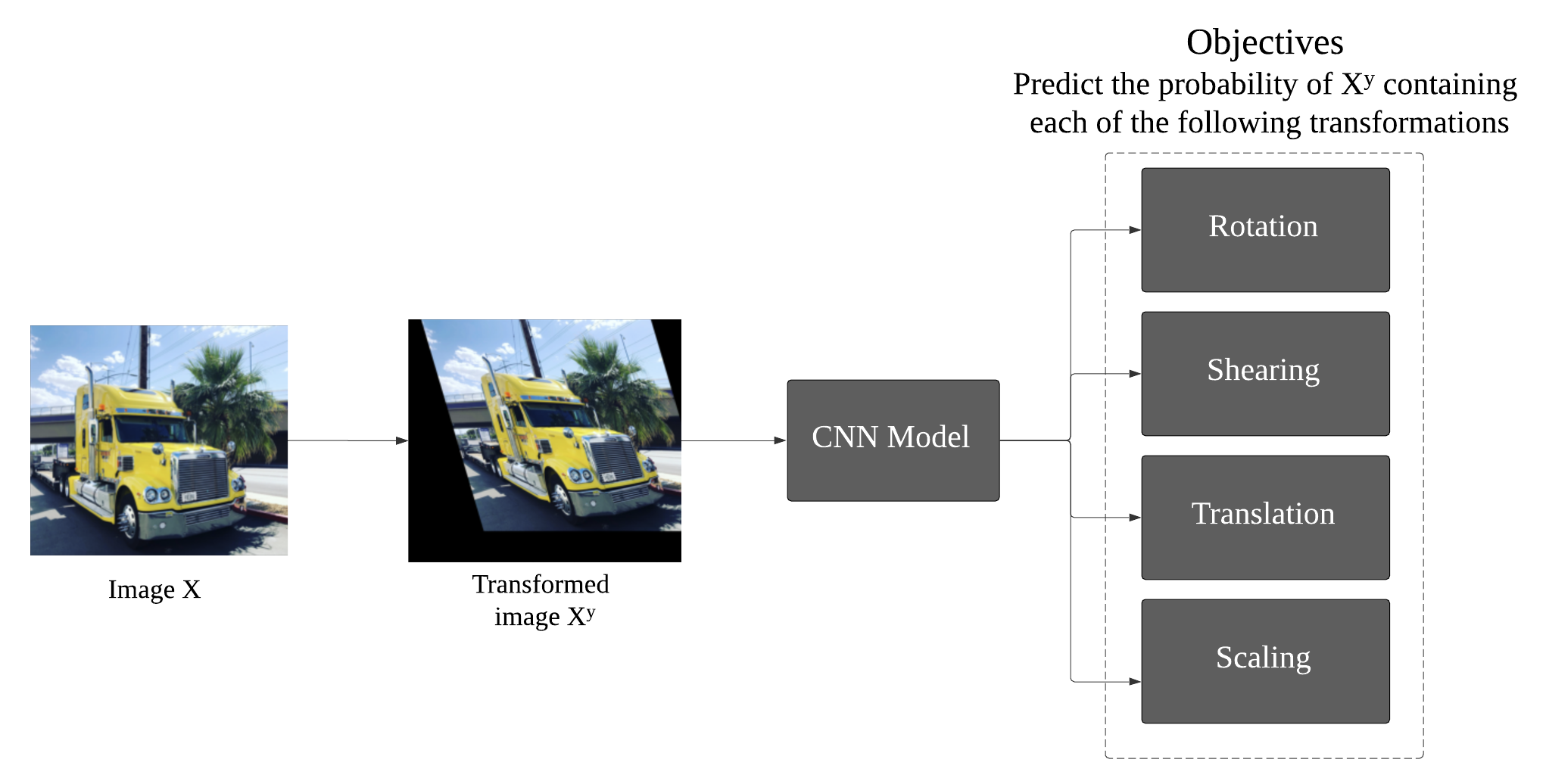}
        \label{fig:illufinal}
\end{figure}

In the second part of the project, we explored more affine transformations, specifically, rotation, translation, shearing and scaling. We applied 5 and 10 transformations in different settings to each image. Specifically, we improvised the pretext task as the prediction of all the transformations applied to the input image, as shown in Figure \ref{fig:illufinal}

\subsubsection{Translation}
It involves shifting images left, right, up, or down. Learning to recognise this transformation can force the CNN to not learn positional bias in the data \cite{aug3}. After the original picture is translated in a direction, the remaining area can be filled with either a constant pixel value such as 0s or 255s, or it can also be filled with random or Gaussian noise. This padding will help in preserving the spatial dimensions of the image after the augmentation. In our method, we fill the remaining space with 0s (black colour).

\subsubsection{Shearing}
Shearing is linear transformation that displaces each point in a fixed orientation, by an amount proportional to its signed distance from the line which is parallel to that orientation and passes through the origin \cite{shearing}. In the process of learning to predict the shearing applied to images, the network is forced to learn angles of relevant features and thus represent the images better. In our experiments, we used shearing factor of 0.3 and -0.3.

\subsubsection{Scaling}
Scaling is a linear transformation that increases (enlarges) or diminishes (shrinks) objects by a scale factor which is the same in all orientations \cite{scaling}. To be able to recognise if an image is scaled, the network would have to learn the size of relevant features in the image. If the value of scaling factor is set to less than 1, we decrease the size of the objects in the image \cite{scaling2}. If the value scaling factor is set to more than 1, we increase size of the object in the image. We used scaling factors of 0.7 and 1.3.

\begin{figure}[h]

\begin{subfigure}{0.195\textwidth}
    \includegraphics[width=0.9\linewidth, height=0.9\linewidth]{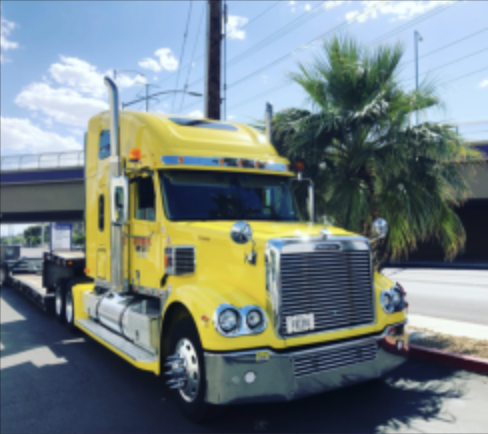} 
    \caption{Original}
    \label{fig:itori}
\end{subfigure}
\begin{subfigure}{0.195\textwidth}
    \includegraphics[width=0.9\linewidth, height=0.9\linewidth]{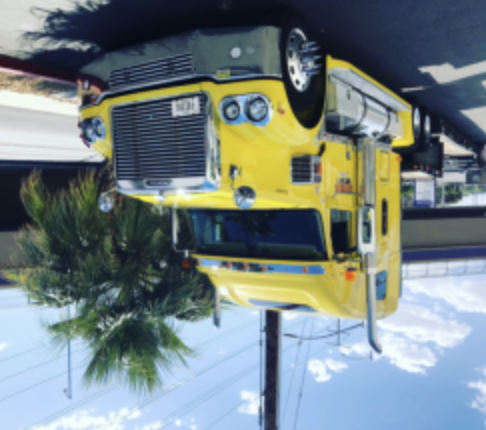}
    \caption{Rotation}
    \label{fig:itrot}
\end{subfigure}
\begin{subfigure}{0.195\textwidth}
    \includegraphics[width=0.9\linewidth, height=0.9\linewidth]{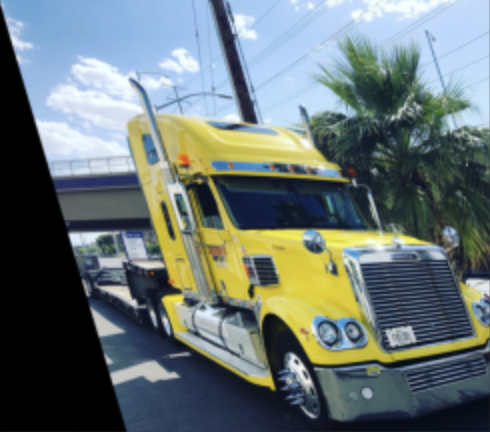}
    \caption{Shearing}
    \label{fig:itshe}
\end{subfigure}
\begin{subfigure}{0.195\textwidth}
    \includegraphics[width=0.9\linewidth, height=0.9\linewidth]{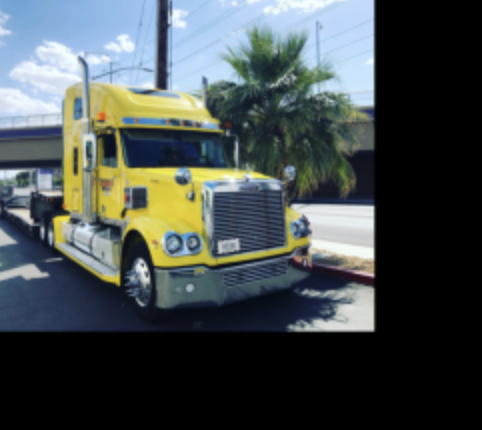}
    \caption{Scaling}
    \label{fig:itsca}
\end{subfigure}
\begin{subfigure}{0.195\textwidth}
    \includegraphics[width=0.9\linewidth, height=0.9\linewidth]{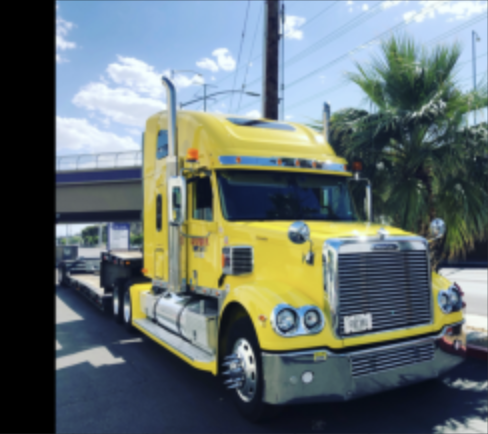}
    \caption{Translation}
    \label{fig:ittra}
\end{subfigure}

\caption{Illustration of various transformations}
\centering
\label{fig:illutrans}
\end{figure}

\subsection{Data augmentation techniques}
Here, we will describe some of the data augmentation techniques that we used in an experiment in the second phase of our project. 

\textbf{Horizontal flipping}: It involves flipping an image about the horizontal axis and is a lot more common than flipping the vertical axis \cite{aug3}. It is easy to implement and has shown to be useful on datasets such as CIFAR-10 and ImageNet. 

\textbf{Random brightness}: The RGB values are easily manipulated using trivial matrix operations in order to increase or reduce the brightness of the image. In our method for data augmentation, brightness of each image is changed to a random value within a specified range.

\textbf{Random zooming}: In this method for data augmentation, images are randomly zoomed in or out in a specified range \cite{aug}. When zoomed in, new pixel values are either added  around the image or pixel values are interpolated respectively \cite{aug2}.

\section{Experiments}
\subsection{Phase I: Rotation prediction}
In the experiments in the first phase of our project, we implemented multiple models and trained them on the pretext task of rotation prediction . We used RMSprop with a batch size of 128, rho of 0.9, and learning rate (lr) of 0.001. We dropped lr by a factor of five after thirty, sixty and eighty epochs. We trained the models for 100 epochs and fed it all four rotated copies of each image \cite{ssl_code}. 

We conducted extensive experiments to test the performance of various models with varied number of rotations. To investigate how much the depth of the models affect the standard of features that are learned, we trained models with different number of convolutional blocks in the architectures. On top of that we also assess the models' performance when the initial blocks are frozen and not trained in the downstream task versus when they are not frozen and are trained to fine tune in the downstream task. While the former measures the models' feature extracting accuracy, the latter measures the fine tuning accuracy. 

We trained classifiers above the feature maps produced by each convolutional block of each model for 50 epochs each for frozen and unfrozen setting. These classifiers are trained in a supervised manner on CIFAR-10 object recognition task. They comprise of 3 fully connected layers; 2 hidden layers comprise of 200 feature channels each and are followed by batch-normalisation and relu units. 

\subsection{Phase II}
In the second phase of our project, we conducted three sets of experiments to test the effect of more transformations, data augmentation and optimizers respectively.

\subsubsection{Affine transformation prediction}
In this set of experiments, we implemented the reformulated problem of affine transformation prediction. We trained the models with 5 and 10 transformations. All the other settings were same as those in the first phase of our project.

\subsubsection{Effect of data augmentation}
Another experiment we conducted was to identify the effect of data augmentation in training in the downstream task of object recognition on CIFAR-10. For this purpose, we chose the best performing model, VGG16 with 2 convolutional blocks. Then, we experimented with no, weak (less) and strong (more) augmentations. Weak augmentation included random zooming in range (0.5, 1), width shift in range (-2, 2), height shift up to 0.1 and horizontal flip. Strong augmentation included those techniques in weak augmentation as well as random rotation up to 45 degrees and random brightness change in range (0.5, 1).

\subsubsection{Search for the best optimizer}
In order to choose the best optimizer, we conducted an experiment by training the best performing model, VGG16 with 2 convolutional blocks, in the downstream task of object recognition task on CIFAR-10 with 3 different optimizers, namely Adam, RMSprop and SGD.

\section{Results}
\subsection{Phase I}

\begin{figure}[h]
    
    \caption{Change in accuracy as training progressed during object recognition downstream task for VGG16 model trained with 2 convolutional blocks for the rotation prediction task with 2, 4 and 8 rotations. Here, fine tuning refers to unfrozen setting and feature extracting refers to frozen setting.}
    \centering
    \includegraphics[height=9cm, width=12cm]{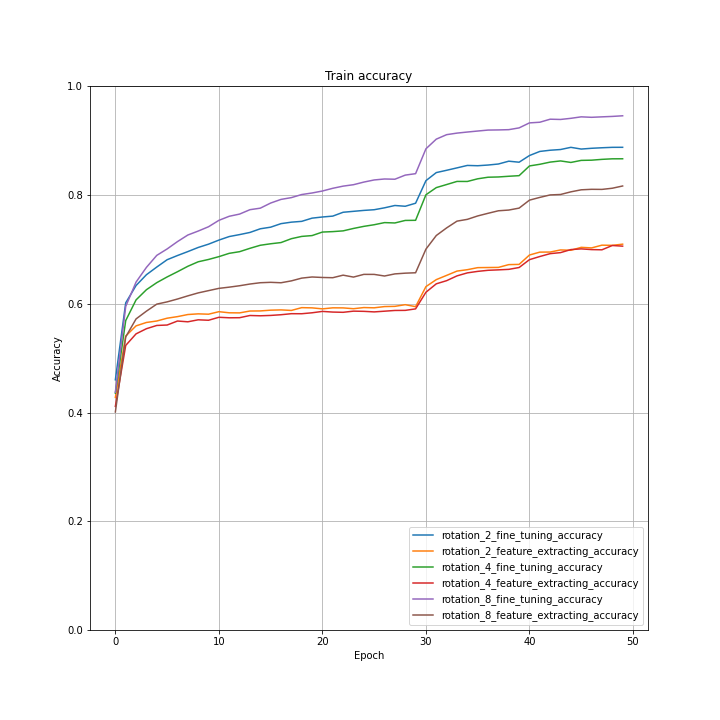} 
    
    \label{fig:vgg}
\end{figure}

The experiments produced interesting results, as shown in Table \ref{tab:resultsfirst}. VGG16 model trained with 2 convolutional blocks for the rotation prediction task with 2 rotations (0 and 180 degrees) for the pretext produced best results in the downstream task of object recognition on CIFAR-10 with almost 80 \% accuracy when the layers trained in pretext task are not retrained in the downstream task and 70\% accuracy when those layers are retrained. Figure \ref{fig:vgg} shows the change in accuracy as training progressed during the downstream task for the model mentioned above.

Overall, the rotation prediction pretext task where the input images were rotated 2 times (0 and 180 degrees) performed better in the downstream task. Also, using less convolutional blocks generally performed better. Using more convolutional blocks produces a gradual decrease in object recognition accuracy, which we believe is because the feature learnt in these layers start to become more specific on the rotation prediction task. Furthermore, we see that increased depth of models lead to better performance in object recognition with regards to the feature maps produced by earlier layers. We believe this is because a deeper model enables the features of layers early on to be less peculiar to the rotation prediction task.

\begin{table}[htbp]
\caption{Classification accuracy in the downstream task of object recognition on CIFAR-10 test set. Models used are mentioned along with the number of used convolutional blocks in the models. Accuracy is reported with varying number of transformations in the pretext task. Frozen refers to the setting where the model is trained on transformation prediction pretext task, then the layers are kept frozen and not trained when non-linear layers are added on top of them which are trained during the training of downstream task of object recognition. Unfrozen refers to the setting where the layers trained in the pretext task are retrained during training of downstream task}

\begin{subtable}{\linewidth}\centering
\caption{Phase I: Rotation prediction}
\begin{tabular}{@{}c|cccccc@{}}
\toprule
\textbf{Model - block}        & \multicolumn{2}{c}{\textbf{Rotation - 2}}                                     & \multicolumn{2}{c}{\textbf{Rotation - 4}}                                     & \multicolumn{2}{c}{\textbf{Rotation - 8}}                                     \\ \midrule
\multicolumn{1}{l|}{\textbf{}} & \multicolumn{1}{l}{\textbf{Unfrozen}} & \multicolumn{1}{l}{\textbf{Frozen}} & \multicolumn{1}{l}{\textbf{Unfrozen}} & \multicolumn{1}{l}{\textbf{Frozen}} & \multicolumn{1}{l}{\textbf{Unfrozen}} & \multicolumn{1}{l}{\textbf{Frozen}} \\ \midrule
ResNet50 - 2                  & 0.7179                                 & 0.6447                               & 0.6145                                 & 0.5809                               & 0.6064                                 & 0.5924                               \\
ResNet50 - 3                  & 0.7084                                 & 0.5807                               & 0.6563                                 & 0.5897                               & 0.6306                                 & \textbf{0.6406}                               \\
ResNet50 - 4                  & 0.6927                                 & 0.3156                               & 0.6446                                 & 0.3604                               & 0.6284                                 & 0.3466                               \\
ResNet50 - 5                  & 0.672                                  & 0.1692                               & 0.6455                                 & 0.1655                               & 0.6024                                 & 0.1678                               \\
DenseNet201 - 2                   & 0.712                                  & 0.5444                               & 0.6639                                 & 0.5249                               & 0.6112                                 & 0.5094                               \\
DenseNet201 - 4                   & 0.7341                                 & 0.1824                               & 0.7147                                 & 0.3899                               & 0.6704                                 & 0.3888                               \\
VGG16 - 2                      & \textbf{0.8083}                                 & \textbf{0.6957}                               & \textbf{0.7397}                                 & \textbf{0.6647}                               & \textbf{0.7062}                                 & 0.6344                               \\
VGG16 - 5                      & 0.1024                                 & 0.0972                               & 0.0964                                 & 0.0964                               & 0.0824                                 & 0.0976                               \\
NIN - 2                           & 0.7153                                 & 0.502                                & 0.649                                  & 0.5036                               & 0.6148                                 & 0.4906                               \\
ResNet152V2 - 2                  & 0.6979                                 & 0.6788                               & 0.622                                  & 0.6559                               & 0.5796                                 & 0.652                                \\ \bottomrule
\end{tabular}
\label{tab:resultsfirst}
\end{subtable}

\begin{subtable}{\linewidth}\centering
\caption{Phase II: Affine transformation prediction}
\begin{tabular}{@{}c|cccc@{}}
\toprule
\textbf{Model - block} & \multicolumn{2}{c}{\textbf{Transform - 5}}                                    & \multicolumn{2}{c}{\textbf{Transform - 10}}                                   \\ \midrule
\textbf{}              & \multicolumn{1}{l}{\textbf{Unfrozen}} & \multicolumn{1}{l}{\textbf{Frozen}} & \multicolumn{1}{l}{\textbf{Unfrozen}} & \multicolumn{1}{l}{\textbf{Frozen}} \\ \midrule
NIN - 1                & 0.7357                                 & 0.5196                               & 0.7319                                 & 0.5538                               \\
VGG16 - 1              & \textbf{0.7672}                        & \textbf{0.6319}                      & \textbf{0.7605}                        & \textbf{0.6338}                      \\ \midrule
ResNet50 - 2           & 0.7633                                 & 0.5006                               & 0.7443                                 & 0.4934                               \\
ResNet152V2 - 2        & 0.7312                                 & 0.4853                               & 0.7188                                 & 0.4419                               \\
NIN - 2                & 0.7494                                 & 0.4385                               & 0.7537                                 & 0.4742                               \\
DenseNet - 2           & 0.7401                                 & 0.475                                & 0.7371                                 & 0.4818                               \\
VGG16 -2               & \textbf{0.8216}                        & \textbf{0.5177}                      & \textbf{0.8291}                        & \textbf{0.5952}                      \\ \midrule
ResNet50 - 3           & 0.6693                                 & 0.3729                               & 0.6467                                 & 0.3893                               \\
ResNet152V2 - 3        & 0.7111                                 & 0.332                                & 0.649                                  & 0.3398                               \\
DenseNet201 - 3        & 0.7256                                 & \textbf{0.3731}                      & 0.7193                                 & \textbf{0.3929}                      \\
VGG16 - 3              & \textbf{0.8323}                        & 0.3596                               & \textbf{0.8068}                        & 0.3855                               \\ \midrule
ResNet50 - 4           & 0.6404                                 & \textbf{0.252}                       & 0.6108                                 & \textbf{0.2837}                      \\
ResNet152V2 - 4        & 0.6885                                 & 0.2321                               & 0.6251                                 & 0.2352                               \\
DenseNet201 - 4        & 0.7054                                 & 0.2462                               & 0.6985                                 & 0.2546                               \\
VGG16 - 4              & \textbf{0.8048}                        & 0.2334                               & \textbf{0.7698}                        & 0.2807                               \\ \midrule
ResNet50 - 5           & 0.6312                                 & 0.1628                               & 0.6013                                 & 0.1885                               \\
ResNet152V2 - 5        & 0.6785                                 & 0.1932                               & 0.6269                                 & 0.1987                               \\
DenseNet201 - 5        & 0.7069                                 & \textbf{0.2447}                      & 0.689                                  & \textbf{0.2532}                      \\
VGG16 - 5              & \textbf{0.7947}                        & 0.1688                               & \textbf{0.7648}                        & 0.2289                               \\ \bottomrule
\end{tabular}
\label{tab:resultssecond}
\end{subtable}

\begin{subtable}{\linewidth}\centering
\caption{Phase II: Effect of data augmentation}
\begin{tabular}{@{}c|cccccc@{}}
\toprule
\textbf{VGG16} & \multicolumn{3}{c}{\textbf{Unfrozen}}              & \multicolumn{3}{c}{\textbf{Frozen}}                \\ \midrule
\textbf{Conv2} & \textbf{Rot-2}  & \textbf{Rot-4}  & \textbf{Rot-8}  & \textbf{Rot-2}  & \textbf{Rot-4}  & \textbf{Rot-8}  \\ \midrule
Strong         & 0.706           & 0.7275          & 0.7171          & 0.5479          & 0.5828          & 0.5569          \\
Weak           & 0.7818          & 0.7946          & 0.7768          & 0.6033          & 0.6236          & 0.607           \\
None           & \textbf{0.8421} & \textbf{0.8544} & \textbf{0.8488} & \textbf{0.7249} & \textbf{0.7541} & \textbf{0.7467} \\ \bottomrule
\end{tabular}
\label{tab:resultssecondaug}
\end{subtable}
\end{table}

\pagebreak

\begin{table}[htbp]
\ContinuedFloat
\begin{subtable}{\linewidth}\centering
\caption{Phase II: Search for the best optimizer}
\begin{tabular}{@{}c|cccccc@{}}
\toprule
\textbf{VGG16} & \multicolumn{3}{c}{\textbf{Unfrozen}}              & \multicolumn{3}{c}{\textbf{Frozen}}                \\ \midrule
\textbf{Conv2} & \textbf{Rot-2}  & \textbf{Rot-4}  & \textbf{Rot-8}  & \textbf{Rot-2}  & \textbf{Rot-4}  & \textbf{Rot-8}  \\ \midrule
SGD            & 0.7524          & 0.5399          & 0.7631          & 0.7038          & 0.7409          & 0.7242          \\
RMSprop        & \textbf{0.8409} & 0.8541          & 0.8466          & 0.7229          & 0.7569          & 0.7412          \\
Adam           & 0.8404          & \textbf{0.8556} & \textbf{0.8487} & \textbf{0.7301} & \textbf{0.7644} & \textbf{0.7477} \\ \bottomrule
\end{tabular}
\label{tab:resultssecondopt}
\end{subtable}

\label{tab:results}
\end{table}

\subsection{Phase II}
All sets of experiments in the second phase presented fascinating results as shown in Table \ref{tab:resultssecond}. Generally, transformation prediction task produced better models in the unfrozen setting, where the layers trained in the pretext task are retrained during training of downstream task. VGG16 model with 3 convolutional blocks performed the best in this setting, achieving accuracy of 83.23\%. But in frozen settings, here the layers trained in the pretext task are frozen and not retrained during training of downstream task, rotation prediction task produced better models. 

\begin{figure}[h]
    
    \caption{Change in accuracy as training progressed during object recognition downstream task for VGG16 model trained in the pretext transformation prediction task with 5 transformations}
    \centering
    \includegraphics[height=9cm, width=12cm]{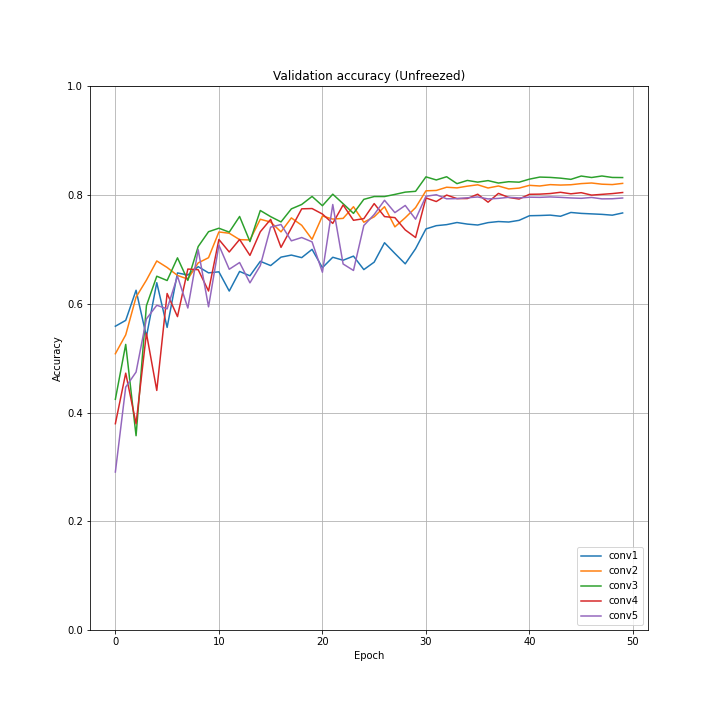} 
    
    \label{fig:transvgg}
\end{figure}

When it comes to the variation in convolutional blocks, trends similar to those in rotation prediction task were observed here as well. Figure \ref{fig:transvgg} shows that VGG16 model trained with 3 convolutional block performs best. The reason for that could be that 3 convolutional blocks provides the network the capacity to learn hierarchical features and at the same time does not make it learn features specific to the pretext task. This trend of 3 convolutional blocks performing well can be seen across models.

The second set of experiments in phase II indicated that data augmentation could not help improve accuracy in the downstream task, as shown in Table \ref{tab:resultssecondaug}. In fact, applying augmentations led to decrease in model accuracy. This could be due to the model not having enough capacity, or because the models in downstream task are fine tuned instead of training from scratch. The search for the best optimizer showed that Adam optimized the models best, with RMSprop not much far off, as shown in Table \ref{tab:resultssecondopt}.

\section{Conclusion}

We began with the aim of solving the problem of scarcity of training data. We come across many works that showed that geometric transformations are powerful supervisory signals for self supervised learning \cite{rotation, aet}. This encouraged us to explore this domain. We then formulated the problem as geometric transformation prediction, and experimented with rotations in phase I. We then expanded to more geometric transformations such as shearing, scaling and translation.

We experimented with multiple models for the self-supervised feature learning task of training a CNN model to predict the image rotation as well as other transformations applied to its input images.
It could be seen that tasks of rotation and transformation prediction forces the CNN model trained on it to learn meaningful features useful for object recognition task as demonstrated. We ran extensive series of  experiment to measure various CNN model's performance on the mentioned self-supervised tasks and evaluated their performance on the downstream task of object recognition on CIFAR-10. It is observed that transformation prediction task produced better models in the unfrozen setting. \\
\\VGG16 model with 3 convolutional blocks exceeded all other models with Top-1 accuracy of 83.23\% in unfrozen setting. In frozen setting, rotation prediction task performed well. Similarly application of complex data augmentation techniques did not help improve the accuracy. As inherently Adam combines the best properties of RMSProp and other optimizers, it optimized well in comparison to other optimizers. So on the basis of qualitative and quantitative analysis we can say that  using VGG16 with 3-conv blocks , as well as using Adam optimizer in our model give us much better results in comparison to other models. 

We can conclude by emphasizing that our model which comprised of SSL with geometric transformations is forced by pretext task to learn good representations so that they contain sufficient information about visual structures of  transformed images. We further demonstrate that a huge number of transformations can be easily included into this framework and the experiment results show us sufficient increase in performance when we transfer our unsupervised learned features on our downstream task of  CIFAR-10 classification.

\clearpage
\bibliographystyle{abbrv}
\bibliography{sample.bib}






\end{document}